\def\year{2019}\relax
\documentclass[letterpaper]{article} 
\usepackage{aaai19}  
\usepackage{times}  
\usepackage{helvet}  
\usepackage{courier}  
\usepackage{url}  
\usepackage{graphicx}  
\usepackage[english]{babel} 
\usepackage[utf8x]{inputenc}
\usepackage[T1]{fontenc}
\usepackage{mathrsfs}
\usepackage{amssymb}
\usepackage{amsmath}
\usepackage{algorithm}
\usepackage{algpseudocode}

\usepackage{graphics}
\usepackage{epsfig}
\usepackage{enumerate}
\usepackage{subfigure}
\usepackage{float}
\usepackage{caption}
\usepackage{booktabs} 

\usepackage{multirow}  

\newtheorem{theorem}{Theorem}

\newtheorem{definition}{Definition}
\newtheorem{proposition}{Proposition}

\begin{document}
	\title{Capacity Control of ReLU Neural Networks by Basis-path Norm}
	\author{Shuxin Zheng$^{1,}$\thanks{This work was done when the author was visiting Microsoft Research Asia.}, Qi Meng$^2$, Huishuai Zhang$^2$, Wei Chen$^2$, Nenghai Yu$^1$, \and Tie-Yan Liu$^2$ \\
		$^1$University of Science and Technology of China\\
		$^2$Microsoft Research Asia \\
		zhengsx@mail.ustc.edu.cn, \{meq, huzhang, wche, Tie-Yan.Liu\}@microsoft.com, ynh@ustc.edu.cn
	}
	\maketitle
\begin{abstract}
Recently, path norm was proposed as a new capacity measure for neural networks with Rectified Linear Unit (ReLU) activation function, which takes the rescaling-invariant property of ReLU into account. It has been shown that the generalization error bound in terms of the path norm explains the empirical generalization behaviors of the ReLU neural networks better than that of other capacity measures. Moreover, optimization algorithms which take path norm as the regularization term to the loss function, like Path-SGD, have been shown to achieve better generalization performance. However, the path norm counts the values of all paths, and hence the capacity measure based on path norm could be improperly influenced by the dependency among different paths. It is also known that each path of a ReLU network can be represented by a small group of linearly independent basis paths with multiplication and division operation, which indicates that the generalization behavior of the network only depends on only a few basis paths. Motivated by this, we propose a new norm \emph{Basis-path Norm} based on a group of linearly independent paths to measure the capacity of neural networks more accurately. We establish a generalization error bound based on this basis path norm, and show it explains the generalization behaviors of ReLU networks more accurately than previous capacity measures via extensive experiments. In addition, we develop optimization algorithms which minimize the empirical risk regularized by the basis-path norm. Our experiments on benchmark datasets demonstrate that the proposed regularization method achieves clearly better performance on the test set than the previous regularization approaches.

\end{abstract}
	
\section{Introduction}
Deep neural networks have pushed the frontiers of a wide variety of AI tasks in recent years such as speech recognition \cite{xiong2016achieving,chan2016listen}, computer vision \cite{ioffe2015batch,ren2015faster} and neural language processing \cite{bahdanau2014neural,gehring2017convolutional}, etc. More surprisingly, deep neural networks generalize well, even when the number of parameters is significantly larger than the amount of training data \cite{zhang2016understanding}. To explain the generalization ability of neural networks, researchers commonly used different norms of network parameters to measure the capacity \cite{bartlett2017spectrally,neyshabur2014search,neyshabur2015norm}.

Among different types of deep neural networks, ReLU networks (i.e., neural networks with ReLU activations \cite{glorot2011deep}) have demonstrated their outstanding performances in many fields such as image classification \cite{resnet,densenet}, information system \cite{cheng2016wide,he2017neural}, and text understanding \cite{t2t} etc. It is well known that ReLU neural networks are positively scale invariant \cite{pathsgd,pathsgd-rnn}. That is, for a hidden node with ReLU activation, if all of its incoming weights are multiplied by a positive constant $c$ and its outgoing weights are divided by the same constant, the neural network with the new weights will generate exactly the same output as the old one for any arbitrary input. \cite{pathsgd} considered the product of weights along all paths from the input to output units as path norm which is invariant to the rescaling of weights, and proposed Path-SGD which takes path norm as the regularization term to the loss function.

\begin{figure}[t]
	\centering
	\includegraphics[width=0.2\textwidth]{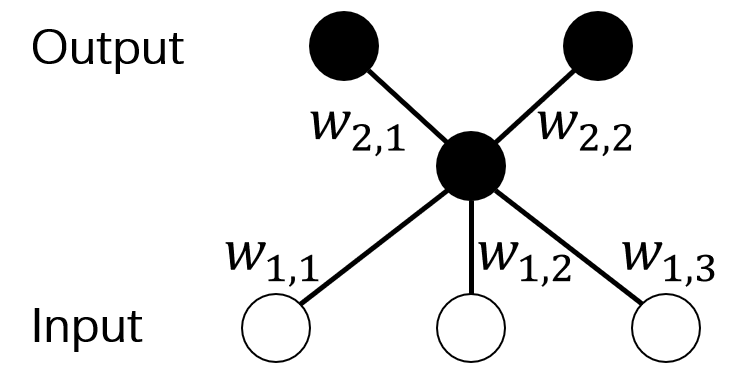}
	\caption{A toy neural network example. The network has 6 paths $p_{i,j}$, where $i \in \{1,2,3\}$ and $j \in \{1,2\}$, the values of paths $v_{p_{i,j}}=w_{1,i}w_{2,j}$,  We can see the dependency among the paths, i.e., $v_{p_{2,2}}=\frac{v_{p_{1,2}}\cdot v_{p_{2,1}}}{v_{p_{1,1}}}$ and $v_{p_{3,2}}=\frac{v_{p_{3,1}}\cdot v_{p_{1,2}}}{v_{p_{1,1}}}$. In this group of basis paths, $p_{1,1}$ is the negative Basis-path, $p_{1,2}$,$p_{2,1}$ and $p_{3,1}$ are the positive basis paths. 
	}\label{fig:sk}
	\vspace{-1.em}
\end{figure}

In fact, each path in a ReLU network can be represented by a small group of generalized linearly independent paths (we call them \emph{basis-path} in the sequels) with multiplication and division operation as shown in Figure \ref{fig:sk}. Thus, there is dependency among different paths. The smaller the percentage of basis paths, the higher the dependency. As the network is determined only by the basis paths, the generalization property of the network should depend only on the basis paths, as well as the relevant regularization methods. In addition, Path-SGD controls the capacity by solving argmin of the regularized loss function, the solution of the argmin problem is approximate because dependency among different values of all paths is not considered in the network. This motivates us to establish a capacity bound based on only the basis paths instead of all the paths. This is in contrast to the generalization bound based on the path norm which counts the values of all the paths and does not consider the dependency among different paths. To tackle these problems, we define a new norm based on the values of the basis paths called \emph{Basis-path Norm}. In previous work, \cite{ecsgd} constructed the basis paths by \emph{skeleton method} and proved that the values of all other paths can be calculated using the values of basis paths by multiplication and division operations. In this work, we take one step further and categorize the basis paths into \emph{positive} and \emph{negative} basis paths according to the sign of their coefficients in the calculations of non-basis paths. 

In order to control generalization error, we need to keep the hypothesis space being small. As we know, loss function can be computed by paths, hence we keep the values of all paths being small. To keep small values of non-basis paths represented by positive and negative basis paths, we control the positive basis paths not being too large while the negative basis paths not being too small. In addition, to keep small values of basis paths, we control the negative basis paths not being too large as well. With this consideration, we define the new Basis-path norm. We prove a generalization error bound for ReLU networks in terms of the basis-path norm. We then study the relationship between this basis-path norm bound and the empirical generalization gap – the absolute difference between test error and training error. The experiments included ReLU networks with different depths, widths, and levels of randomness to the label. For comparison purpose, we also compute the generalization error bounds induced by other capacity measures for neural networks proposed in the literature. Our experiments show that the generalization bound based on basis-path norm is much more consistent with the empirical generalization gap than those based on other norms. In particular, when the network size is small, the ordinary path norm bound fit empirical generalization gap well. However, when the width and depth increases, the percentage of non-basis paths increases, and the dependency among paths increases and we observe that the path norm bound degenerates in reflecting the empirical generalization gap. In contrast, our basis-path norm bound fits the empirical generalization gap consistently as the network size changes. This validates the efficacy of BP norm as a capacity measure. 

Finally, we propose a novel regularization method, called Basis-path regularization (BP regularization), in which we penalize the loss function by the BP norm. Empirically, we first conduct experiments on recommendation system of MovieLens-1M dataset to compare the multi-layer perceptron (MLP) model's generalization with BP regularization and baseline norm-based regularization, then we verify the effectiveness of BP regularization on image classification task with ResNet and PlainNet on CIFAR-10 dataset. The results of all experiments show that, with our method, optimization algorithms (i.e., SGD, Adam, Quotient SGD) can attain better test accuracy than with other regularization methods.

\subsection{Related Work}
Generalization of deep neural networks has attracted a great deal of attention in the community \cite{zhang2016understanding,neyshabur2017exploring,kawaguchi2017generalization}. Norm and margin-based measures have been widely studied, and commonly used in neural network optimization with capacity control \cite{bartlett2002rademacher,evgeniou2000regularization,neyshabur2015norm}. For example, in \cite{bartlett2017spectrally}, the authors proposed a margin-based generalization bound for networks that scale with their margin-normalized spectral complexity. An analysis of generalization bounds based on PAC-Bayes was proposed in \cite{dziugaite2017computing}. 

Among these measures, the generalization bound based on path norm is tighter theoretically \cite{neyshabur2015norm}. Empirically, path norm has been showed to be more accurate to describe the tendency of generalization error \cite{neyshabur2017exploring}. Thus, we are interested in the capacity measure which is related to the path norm. In \cite{neyshabur2015norm}, the authors first proposed group norm and path norm. The results show that the path norm is equivalent to a kind of group norm.  In \cite{pathsgd,pathsgd-rnn}, the authors proposed to use path norm as a regularization term for ReLU multi-layers perceptron (MLP) network and recurrent network and designed Path-SGD algorithm. In \cite{neyshabur2017exploring}, the authors empirically compared different kinds of capacity measures including path norm for deep neural network generalization. However, none of those norms considered the dependency among paths in the networks.

\section{Preliminaries}\label{sec2.1}
In this section, we introduce ReLU neural networks and generalization error.

First of all, we briefly introduce the structure of rectifier neural network models. Suppose {\small$f_w: \mathcal{X} \to\ \mathcal{Y}$} is a $L$-layer neural network with weight {\small$w\in\mathcal{W}$}, where input space {\small$\mathcal{X}\subset \mathbb{R}^d$} and output space {\small$\mathcal{Y} = \mathbb{R}^K$}. In the $l$-th layer ($l=0,…,L$), there are $h_l$ nodes. We denote the nodes and their values as {\small$\{O^l,o^l\}$}. It is clear that, $h_0=d, h_L=K$. The layer mapping is given as, $o^l=\sigma(w_l^T o^{l-1})$, where $w_l$ is the adjacency matrix in the $l$-layer, and the rectifier activation function $\sigma(\cdot)=max(\cdot,0)$ is applied element-wisely. We can also calculate the $k$-th output by paths, i.e., {\small \begin{align}
	N_{w}^{k}(x)= \sum_{(i_0,\cdots,i_{L}=k)}\prod_{l=1}^{L}w_l(i_{l-1},i_l)\cdot \prod_{l=1}^{L-1}\mathbb{I}(o^l_{i_{l}}(w,x)>0) \cdot x_{p_0}\label{loss}
	\end{align}}where {\small$(i_0,\cdots,i_L)$} is the path starting from input feature node $O_{i_0}^0$ to output node $O_{i_L}^L$ via hidden nodes {\small$ O^1_{i_1},…,O^{L-1}_{i_{L-1}}$}, and {\small$w_l(i_{l-1},i_l)$} is the weight of the edge connecting nodes {\small$O^{l-1}_{i_{l-1}}$} and {\small$O^{l}_{i_{l}}$}. \footnote{The paths across the bias node can also be described in the same way. For simplicity, we omit the bias term. } 

We denote $p_{(i_0,\cdots,i_L)}=\prod_{l=1}^{L}w_l(i_{l-1},i_l)$ and $a_{(i_0,\cdots,i_{L})}=\prod_{l=1}^{L-1}\mathbb{I}(o^l_{i_{l}}(w,x)>0)$. The output can be represented using paths as
$$N_{p,a}^k(x)=\sum_{(i_0,\cdots,i_L)}p_{(i_0,\cdots,i_L)}\cdot a_{(i_0,\cdots,i_L)}\cdot x_{i_0}.$$
For ease of reference, we omit the explicit index $(i_0,\cdots,i_L)$ and use $i$ be the index of path. We use $p=(p_1,p_2,\cdots,p_{M})$ where $M=\prod_{l=0}^Lh_l$ to denote the path vector. The path norm used in Path-SGD \cite{pathsgd} is defined as $\Omega(p)=\left(\sum_{i=1}^Mp_{i}^2\right)^{1/2}.$ 

Given the training set $\{(x_1,y_1),\cdots,(x_n,y_n)\}$ i.i.d sampled from the underlying distribution $\mathbb{P}$, machine learning algorithms learn a model $f$ from the hypothesis space $\mathcal{F}$ by minimizing the empirical loss function $l(f(x),y)$. The uniform generalization error of empirical risk minimization in hypothesis space $\mathcal{F}$ is defined as
\begin{align*}
\epsilon_{gen}(\mathcal{F})=\sup_{f\in\mathcal{F}}|\frac{1}{n}\sum_{i=1}^nl(f(x_i),y_i)-\mathbb{E}_{(x,y)\sim\mathbb{P}}l(f(x),y)|.
\end{align*} Generalization error $\epsilon_{gen}$ measures how well a model $f$ learned from the training data $S$ can fit an unknown test sample $(x,y)\sim\mathbb{P}$. 

Empirically, we consider the empirical generalization error which is defined as the difference of empirical loss between the training set and test set at the trained model $f$.

\section{Basis-path Norm}\label{sec3}

In this section, we define the Basis-path Norm (abbreviated as BP norm) on the networks. Using the BP norm, we define a capacity measure which is called BP-measure and we prove that the generalization error can be upper bounded using this measure.

\subsection{The Definition of Basis-path Norm}\label{sec3.1}
First, as shown in \cite{ecsgd}, the authors constructed a group of basis paths by skeleton method. It means that the value of non-basis paths can be calculated using the values of basis paths. In the calculation of non-basis paths' values, some basis paths always have positive exponents and hence appear in the numerator, others have negative exponents and hence appear in the denominator. We use $\tilde p$ to denote a non-basis path and $p_1,\cdots,p_r$ to denote basis paths. We have the following proposition.

\begin{proposition}\label{propo1}
	For any non-basis path $\tilde{p}$,  $\tilde{p}=\prod_{i=1}^{m}p_i^{\alpha_i}\prod_{j=m+1}^{r}p_j^{\alpha_j}$, where $\alpha_i\leq 0, \alpha_j\geq 0$.
\end{proposition}
Limited by the space, we put the detailed proof in the supplementary materials. 

The proposition shows that basis paths $p_1,\cdots,p_m$ always have negative exponent in the calculation, while $p_{m+1},\cdots,p^r$ always have positive exponent. We call the basis path with negative exponent $\alpha_i$ \emph{Negative Basis Path} and denote the negative basis path vector as $p^{-}=(p_1,\cdots,p_{m})$. We call the basis path with positive exponent $\alpha_j$ \emph{Positive Basis Path} and denote it as $p^{+}=(p_{m+1},\cdots,p_{r})$.

In order to control generalization error, we need to keep the hypothesis space being small. Thus we want all the paths to have small values. For non-basis path represented by $p_i$ and $p_j$, we control $p_i$ not being too small because $\alpha_i$ is negative, and $p_j$ not being too large because $\alpha_j$ is positive. We control $p_i$ not being too large as well to keep small values of basis paths. We define the following basis-path norm $\phi(\cdot)$ as follows.

\begin{definition}The basis norm on the ReLU networks is  
	\begin{align}\label{eq6}
	\phi(p)=\sup\left\{|\log{|p_1|}|,\cdots,|\log{|p_m|}|, |{p_{m+1}}|,\cdots,|p_{r}|\right\}.
	\end{align}
\end{definition} 

We next provide the property of $\phi(p)$.

\begin{theorem}\label{thm1}
	$\phi(p)$ is a norm in the vector space where $p^-$ is a vector in Euclidean space and $p^+$ is a vector in a generalized linear space under the generalized addition and generalized scalar multiplication operations: $p^-\oplus (p')^-=[p_1\cdot p'_1,\cdots,p_m\cdot p'_m]$ and $c\odot p^-=[\textit{sgn}(p_1)\cdot |p_1|^{a},\cdots,\textit{sgn}(p_m)\cdot |p_m|^{a}]$ for $p^-, (p')^-\in\mathbb{R}^m$ and $c\in\mathbb{R}$.
\end{theorem}

\textit{Proof:} The definition of $\phi(p)$ is equivalent to 	\begin{align}
\phi(p)=\sup\left\{\phi_{\infty}(p^+),\phi_{\infty}(p^-)\right\}.
\end{align}where {\small$\phi_{\infty}(p^-)=\sup_i\left\{|\log{|p_i|}|, i=1,\cdots,m\right\}$} and {\small$\phi_{\infty}(p^+)=\sup_j\left\{|p_j|, j=m+1,\cdots,r\right\}$}. Obviously, $\phi_{\infty}(p^+)$ is the $\ell_{\infty}$ norm in Euclidean space. Thus, it only needs to prove $\phi_{\infty}(p^-)$ is a kind of norm. Next, we prove that $\phi_{\infty}(p^-)$ is a norm in the generalized linear space.

In the generalized linear space, the zero vector is $I$, where $I$ denotes a vector with all elements being equal to $1$. Based on the generalized linear operators, we verify the properties including positive definite, absolutely homogeneous and the triangle inequality of $\|p^-\|_{\infty}$ as follows: 

(1) (Positive definite) 
$\phi_{\infty}(p^-)\geq 0$ and $\phi_{\infty}(p^-)=0$ when $p^-=I$.

(2) (Absolutely homogeneous) For arbitrary $c\in\mathbb{R}$, we have
\begin{align*}
\phi_{\infty}(c\cdot p^-)=\sup_i\left\{|\log |p^-_i|^c|, i=1,\cdots,m\right\}=|c|\cdot\phi_{\infty}(p^-).
\end{align*}

(3) (Triangle inequality)
\begin{align*}
\phi_{\infty}(p^-\oplus (p')^-)=&\sup_i\left\{|\log|p_ip'_i||, i=1,\cdots,m\right\}\\
\leq &\sup_i\left\{|\log|p_i||, i=1,\cdots,m\right\}\\
&+\sup_i\left\{|\log|p'_i||, i=1,\cdots,m\right\}\\
=&\phi_{\infty}(p^-)+\phi_{\infty}((p')^-).
\end{align*} Considered that $\phi_{\infty}(p^+)$ and $\phi_{\infty}(p^-)$ are both norms, taking supreme of them is still a norm. Thus $\phi(p)$ satisfies the definition of norm.   $\Box$

\subsection{Generalization Error Bound by Basis-path Norm}\label{sec3.2}
We want to use the basis-path norm to define a capacity measure to get the upper bound for the generalization error. Suppose the binary classifier is given as $g(x)=v^Tf(x)$, where $v$ represents the linear operator on the output of the deep network with input vector $x\in\mathbb{R}^d$. 
We consider the following hypothesis space which is composed of linear operator $v$, and $L$-layered fully connected neural networks with width $H$ and input dimension $d$:
\begin{align*}
&\mathcal{G}_{\gamma,v}^{d,H,L}=\\
&\{g=v\circ f: L\geq 2, h_0=d, h_1=\cdots=h_{L-1}=H, \phi(p)\leq \gamma\}.
\end{align*}

\begin{theorem}\label{thm3.1}
	Given the training set $\{(x_1,y_1),\cdots,(x_n,y_n)\}$ with $x_i\in\mathbb{R}^d$, $y_i\in\{0,1\}$, and the hypothesis space $\mathcal{G}_{\gamma,v}^{d,H,L}$ which contains MLPs with depth $L\geq 2$, width $H$ and $\phi(p)$, for arbitrary $z>0$, for every $\delta>0$, with probability at least $1-\delta$, for every hypothesis $g\in\mathcal{G}_{\gamma,v}^{d,H,L}$, the generalization error can be upper bounded as
	\begin{align*}
	&\epsilon_{gen}(\mathcal{G}_{\gamma,v}^{d,H,L}) \leq 4\sqrt{\frac{2\ln(4/\delta)}{n}}+\\
	&\quad \quad 2\sqrt{\frac{2\Phi(\gamma; d, H, L)(4H)^{L-1}\cdot\|v\|_2^2\cdot\max_i\|x_i\|_2^2}{n}},
	\end{align*}where
	\begin{align}\label{eqn:epnorm}
	&\Phi(\gamma; d, H, L)\overset{\Delta}{=} \nonumber\\
	&(He^{2\gamma}+(d-1)H\gamma^2)\left(1+(H-1)\gamma^2e^{2\gamma}\right)^{L-2}.
	\end{align}
\end{theorem}
We call $\Phi(\gamma; d, H, L)$ \emph{Basis-path measure}.
Therefore, the generalization error $\epsilon_{gen}(\mathcal{G}_{\gamma,v}^{d,H,L})$ can be upper bounded by a function of  Basis-path measure. 

The proof depends on estimating the value of different types of paths and counting the number of different types of paths. We give the proof sketch of Theorem \ref{thm3.1}.

\textbf{Proof of Theorem \ref{thm3.1}:} 

Step 1: If we denote $\mathcal{F}_{\gamma}=\{f: L\geq 2, h_0=d, h_1=\cdots=h_{L-1}=H, \phi(p)\leq \gamma\}$, the generalization error of a binary classification problem is $$\epsilon_{gen}(\mathcal{G}_{\gamma,v}^{d,H,L})\leq 2\|v\|_2^2 \mathcal{RA}(\mathcal{F}_{\gamma})+4\sqrt{\frac{2\ln(4/\delta)}{n}},$$where $\mathcal{RA}(\cdot)$ denotes the Rademacher complexity of a hypothesis space \cite{wolf2018mathematical}. Following the results of Theorem 1 and Theorem 5 in \cite{neyshabur2015norm} under $p=2$ and $q=\infty$, we have $$\mathcal{RA}(\mathcal{F}_{\gamma})\leq \sqrt{\frac{2\Omega^2(\mathcal{F}_{\gamma})(4H)^{L-1}\max_i\|x_i\|_2^2}{n}},$$ where $\Omega(\mathcal{F}_{\gamma})$ is the maximal path norm of $f, f\in\mathcal{F_{\gamma}}$.

Step 2 (estimating path value): We give $\Omega(\mathcal{F}_{\gamma})$ an upper bound using Basis-path norm. Using $\phi(p)\leq\gamma$, we have $e^{-\gamma}\leq|p_i|\leq e^{\gamma}$ and $|p_j|\leq\gamma$. Then using Proposition \ref{propo1}, we have
\begin{align}
|\tilde p|&\leq \big|\prod_{i=1}^{m}p_i^{\alpha_i}\prod_{j=m+1}^{r}p_j^{\alpha_j}\big|\\
&\leq e^{-\gamma\sum_{i=1}^m\alpha_i}\cdot\gamma^{\sum_{i=m+1}^r\alpha_j},
\end{align}where $\alpha_i\leq 0, \alpha_j\geq 0$.

As shown in skeleton method in \cite{ecsgd} (which can also be referred in supplementary materials), basis paths are constructed according to skeleton weights. Here, we clarify the non-basis paths according to the number of non-skeleton weights it contains. We denote the non-basis path which contains $b$ non-skeleton weights as $\tilde p_b$. The proofs of Proposition \ref{propo1} indicates that for $\tilde p_b, \sum_{i=1}^m\alpha_i=1-b, \sum_{i=m+1}^r\alpha_j=b$. Thus we have $$|\tilde p_b|\leq e^{\gamma(b-1)}\cdot\gamma^{b}.$$

Step 3 (counting the number of different type of paths): Based on the construction of basis paths (refer to the skeleton method in supplementary),  in each hidden layer, there are $H$ skeleton weights and $H(H-1)$ non-skeleton weights. We can get that the number of $\tilde p_b$ in a $L$-layer MLP with width $H$ is $(d-1)HC_{L-2}^{b-1}(H-1)^{b-1}+HC_{L-2}^b(H-1)^b$ if $1\leq b\leq L-2$ and $(d-1)H(H-1)^{L-2}$ if $b=L-1$.

Step 4: We have:
\begin{align}
\Omega^2(\mathcal{F}_{\gamma})=\sum_{i=1}^{m}(p_{i})^2+\sum_{j=m+1}^{r}(p_j)^2+\sum_{b=2}^{L-1}\sum_{k}\tilde{p}_{b,k}^2.
\end{align} The number of negative basis paths is $H$ and $e^{-\gamma}\leq |p_i|\leq e^{\gamma}$, so we have $\sum_{i=1}^mp_{i}^2\leq He^{2\gamma}$, where $m=H$. 
\begin{align}
\begin{split}
&\Omega^2(\mathcal{F}_{\gamma})\\
\leq& He^{2\gamma} +\sum_{b=1}^{L-2}\left((d-1)HC_{L-2}^{b-1}(H-1)^{b-1}+HC_{L-2}^b(H-1)^b\right)\\
&\cdot\left(\gamma^be^{\gamma(b-1)}\right)^2 +(d-1)H(H-1)^{L-2}\cdot\left(\gamma^{L-1}e^{\gamma(L-2)}\right)^2 \\
\leq&( He^{2\gamma}+(d-1)H\gamma^2\sum_{b=0}^{L-2}C_{L-2}^b(H-1)^b\left(\gamma^2e^{2\gamma}\right)^{b}\\
\leq&( He^{2\gamma}+(d-1)H\gamma^2)\left(1+(H-1)\gamma^2e^{2\gamma}\right)^{L-2}\\
=&\Phi(\gamma; d, H, L), \label{ineq10}
\end{split}
\end{align}where Ineq.\ref{ineq10} is established by the calculation of $(1+x)^a$.

$\quad\Box$

Based on the above theorem,  we discuss how $\Phi(\gamma;d,H,L)$ changes as width $H$ and depth $L$. (1) For fixed $\gamma$, $\Phi(\gamma;d,H,L)$ increases exponentially as $L$   and $H$ goes to large. (2) $\Phi(\gamma;d,H,L)$ increases as $\gamma$ increases. If $\gamma$ diminishes to zero, we have $\Phi(\gamma;d,H,L)\rightarrow H$. In this case, the feature directly flow into the output, which means that $f_k(x)=x_{k\mod{d}}$, for $k=1,\cdots, H$. (3) If $\gamma=\mathcal{O}\left( \frac{1}{\sqrt{HL}}\right)$, we have $\Phi(\gamma;d,H,L)\leq \mathcal{O}\left(H+\frac{d}{L}\right)$. It increases linearly  as $d$ and $H$ increase and decreases linearly as $L$ increases.  

\subsection{Empirical Verification}

In the previous section, we derived a BP norm induced generalization error bound for ReLU networks. In this section, we study the relationship between this bound and the empirical generalization gap – the absolute difference between test error and training error with real-data experiments, in comparison with the generalization error bounds given by other capacity measures, including weight norm \cite{evgeniou2000regularization}, path norm \cite{neyshabur2015norm} and spectral norm \cite{bartlett2017spectrally}. We follow the experiment settings in \cite{neyshabur2017exploring}, and extend on our BP norm bound. As shown in the previous section, the BP norm with capacity is proportional to Eqn.\ref{eqn:epnorm} We conduct experiments with multi-layer perceptrons (MLP) with ReLU of different depths, widths, and global minima on MNIST classification task which is optimized by stochastic gradient descent. More details of the training strategies can be found in the supplementary materials. All experiments are averaged over 5 trials if without explicit note.

\begin{figure*}[ht]
	\centering
	\includegraphics[width=0.8\textwidth]{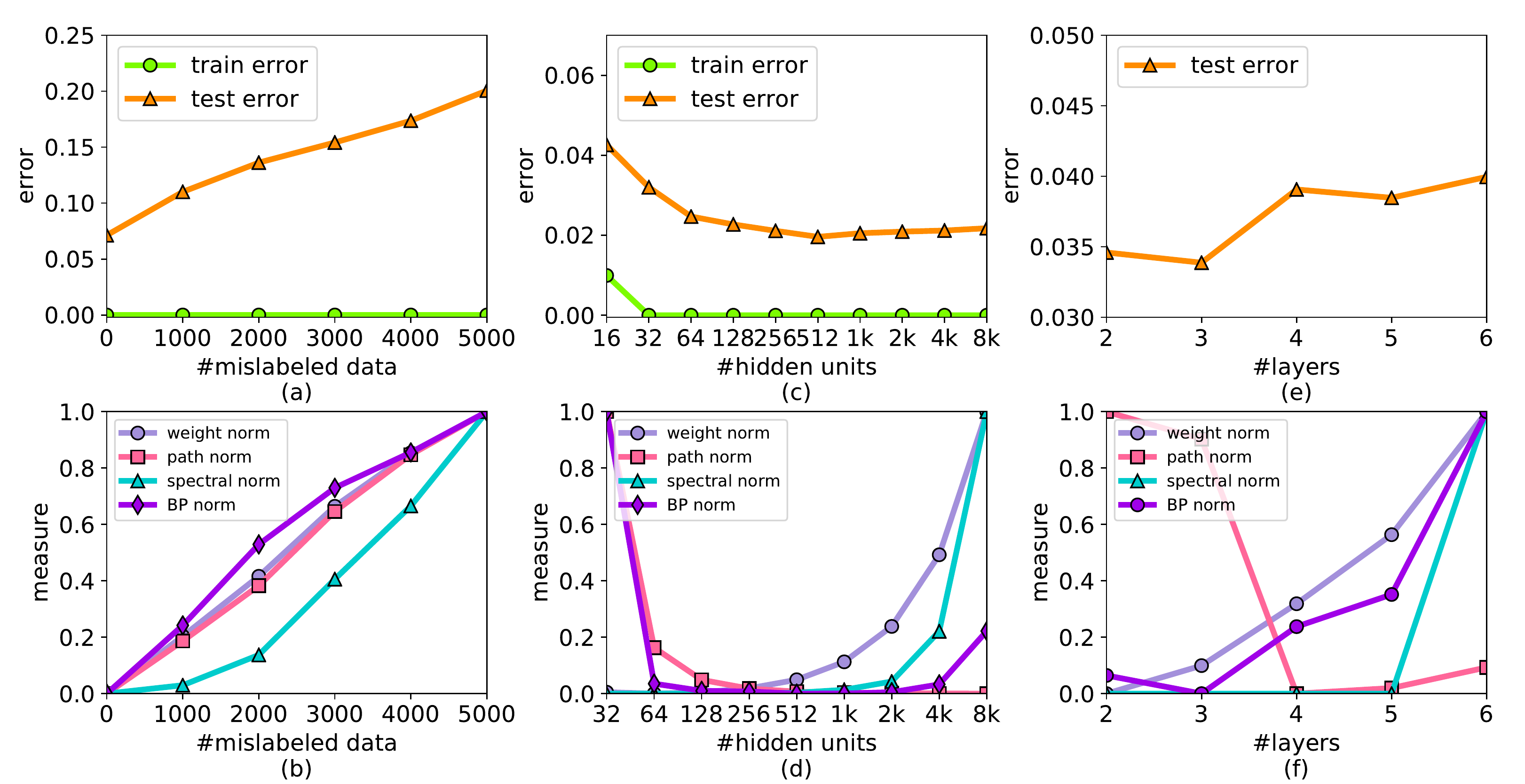}
	\caption{Left: experiments on different global minima for the objective function on the subset with true labels:(a) the training and test error, (b) different measures w.r.t. the size of random labels. Middle: experiments on different hidden units: (c) the training and test error, (d) different measures w.r.t. the size of hidden units for each layer. Right: (e) the test error, (f) different measure w.r.t. the number of layers of the network.}\label{fig:rand_size_depth}
\end{figure*}

First, we train several MLP models and force them to converge to different global minima by intentionally replacing a different number of training data with random labels, and then calculate the capacity measures on these models. The training set consists of 10000 randomly selected samples with true labels and another at most 5000 intentionally mislabeled data which are gradually added into the training set. The evaluation of error rate is conducted on a fixed 10000 validation set. Figure \ref{fig:rand_size_depth} (a) shows that every network is enough to fit the entire training set regardless of the amount of mislabeled data, while the test error of the learned networks increases with increasing size of the mislabeled data. As shown in Figure \ref{fig:rand_size_depth} (b), the measure of BP norm is consistent with the behaviors of generalization on the data and indeed is a good predictor of the generalization error, as well as weight norms, path norm, and spectral norm.

We further investigate the relationship between generalization error and the network size with different widths. We train a bunch of MLPs with 2 hidden layers and varying number of hidden units from 16 to 8192 for each layer. The experiment is conducted on the whole training set with 60000 images. As shown in Figure \ref{fig:rand_size_depth}(c), the networks can fit the whole training set when the number of hidden units is greater than or equal to 32, while the minimal test error is achieved with 512 hidden units, then shows a slightly over fitting on training set beyond 1024 hidden units. Figure \ref{fig:rand_size_depth}(d) shows that the measure of BP norm behaves similarly to the trend of generalization errors which decreases at the beginning and then slightly increases, and also achieves minimal value at 512 hidden units. Weight norm and spectral norm keep increasing along with the network size growing while the trend of generalization error behaves differently. Path norm shows the good explanation of the generalization when the number of hidden units is small, but keeps decreasing along with increasing the network size in this experiment. One possible reason is that the proportion of basis paths in all paths is decreasing, and the vast majority improperly affects the capacity measure when the dependency in the network becomes large. In contrast, BP norm better explains the generalization behaviors regardless of the network size.

Similar empirical observation is shown when we train the network with a different number of hidden layers. Each network has 32 hidden units in each layer and can fit the whole training set in this experiment. As shown in Figure\ref{fig:rand_size_depth}(e,f), the minimal test error is achieved with 3 hidden layers, and then shows an over fitting along with the increasing of the layers. The weight norm keeps increasing with the growing of network size as discussed above, and the $\Pi_i h_i$ in spectral norm will be quite large when layers $L$ is increasing. Path norm can partially explain the decreasing generalization error before 4 hidden layers and it indicates that the networks with 4, 5 and 6 hidden layers have small generalization error, which doesn't match our observations. The amount of non-basis paths is exponentially growing when layers $L$ is increasing, therefore the path norm couldn't measure the capacity accurately by counting all paths' values. In contrast, the BP norm can nearly match the generalization error, these observations verify that BP norm bound is more tight to generalization error and can be a better predictor of generalization.

\section{Basis-path Regularization for ReLU Networks}\label{sec4}

\begin{algorithm}[t]\label{alg:1}
	\caption{Optimize ReLU Network with Basis-path Regularization}
	\label{EC-Opt F}
	\begin{algorithmic} 
		\Require learning rate $\eta_t$, training set $S$, initial $w_0$.
		\For {$t=0,\cdots, T$} 
		\State 1. Draw mini-batch data $x^t$ from $S$.
		\State 2. Compute gradient of the loss function $g(w^t) = \nabla f(w^t,x^t)$.
		\State 3. Compute gradient of the basis-path regularization $h(w^t) = \nabla R(w)$ by Ineq. (\ref{eqn:18}) and (\ref{eq:12}).
		\State 4. Update $w^{t+1} = w^t-\eta_t (g(w^t) + h(w^t))$. 
		\EndFor 
		\Ensure {\small$w^{T}$}.
	\end{algorithmic}
\end{algorithm}

In this section, we propose Basis-path regularization, in which we penalize the loss function by the BP norm. According to the definition of BP norm in Eqn.(\ref{eq6}), to make it small, we need to restrict the values of negative basis paths to be moderate (neither too large nor too small) and minimize the value of positive basis paths. To this end, in our proposed method, we penalize the empirical loss by the $l_2$ distance between the values of negative basis paths and $1$, as well as the sum of the values of all positive basis paths. 

The constraint $\phi(p)\leq \gamma$ equals to $\|p^+\|^2\leq\gamma^2$ and $\|\log(p^-)^2\|^2\leq(2\gamma)^2$, which means that the largest element in a vector is smaller than $\gamma$ iff all of the element is smaller than $\gamma$. We choose to optimize their square because of the smoothness.  So using the Lagrangian dual methods, we add the constraint $\frac{\lambda_1}{2}\|p^+\|^2$ and $\frac{\lambda_2}{4}\|\log(p^-)^2\|^2$ in the loss function and then optimize the regularized empirical risk function:
\begin{align}
\begin{split}
L(w,x) &= f(w,x)+R(p)\\
&=f(w,x)+\frac{\lambda_1}{2}\|p^+\|^2+\frac{\lambda_2}{4}\|\log(p^-)^2\|^2.
\end{split}
\end{align}
We use $g(w)$ to denote the gradient of loss with respect to $w$, i.e., $g(w)=\frac{\partial f(w,x)}{\partial w}$.
For the non-skeleton weight $w_{j}$, since it is contained in only one positive basis path $p_{j}$, we can calculate the gradient of the regularization term with respect to $w_{j}$ as 
{\small\begin{align}\label{eqn:18}h(w_{j})=\frac{\lambda_1}{2}\frac{\partial R(p)}{\partial p_{j}}\frac{\partial p_{j}(w)}{\partial w_{j}}=\lambda_1\cdot\frac{p_{j}^2}{w_{j}}.\end{align}}
For the skeleton weight $w_{i}$, it is contained in only one negative basis path $p_i$ (if the neural network has equal number of hidden nodes) and some of the positive basis paths $p_j$. Thus its gradient can be calculated as follows
\begin{align}\label{eq:12}
\begin{split}
h(w_{i})&=\frac{\lambda_2}{4}\frac{\partial R(p)}{\partial p_{i}}\frac{\partial p_{i}(w)}{\partial w_{i}}+\frac{\lambda_1}{2}\sum_{p_j:w_{i}}\frac{\partial R(p)}{\partial p_{j}}\frac{\partial p_{j}(w)}{\partial w_{i}}\\ &=\frac{\lambda_2\log{p_i}}{w_{i}}+\lambda_1\sum_{p_j:w_{i}}\frac{p_{j}^2}{w_{i}},
\end{split}
\end{align} where $p_j:w_{i}$ denotes all positive basis paths containing $w_i$. 

Combining them together, we get the gradient of the regularized loss function with respected to the weights. For example, if we use stochastic gradient descent to be the optimizer,  the update rule is as follows:
\begin{align}
&w^{t+1}=w^{t}-\eta_t(g(w^t)+ h(w^t)).
\end{align}
Please note that the computation overhead of $h(w_i)$ is high, moreover, we observed that the values of the negative basis paths are relatively stable in the optimization, thus we set $h(w_{i})$ to be zero for ease of the computation. Specifically, basis-path regularization can be easily combined with the optimization algorithm which is in quotient space. 

The flow of SGD with basis-path regularization is shown in Algorithm 1, it's trivial to extend basis-path regularization to other stochastic optimization algorithms. Comparing to weight decay, basis-path regularization has little additional computation overhead. All the additional computations regarding Ineq.(\ref{eqn:18}) only introduce very lightweight element-wise matrix operations, which is small compared with the forward and backward process. 

\begin{figure*}[ht]
	\centering
	\includegraphics[width=1.0\textwidth]{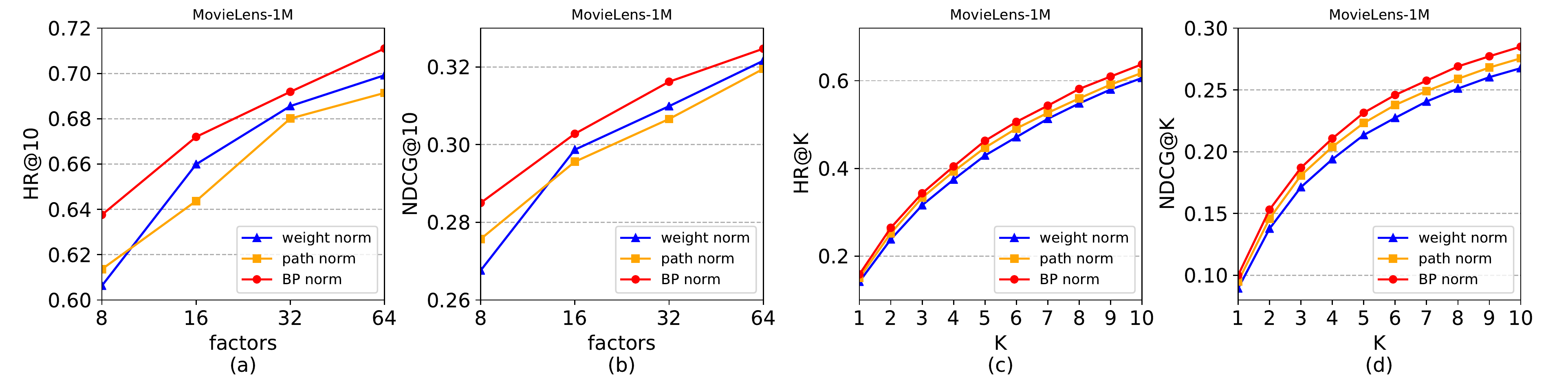}
	\caption{Performance of HR and NDCG w.r.t. the number of predictive factors and Top-K items recommendation.}\label{fig:movielens}
\end{figure*}

\section{Experimental Results}

In this section, we evaluate Basis-path Regularization on deep ReLU neural networks with the aim of verifying that does our proposed BP regularization outperforms other baseline regularization methods and whether it can improve the generalization on the benchmark datasets. For sake of fairness, we reported the \textit{mean} of 5 independent runs with random initialization. 

\subsection{Recommendation System}

We first apply our basis-path regularization method to recommendation task with MLP networks and conduct experimental studies based on a public dataset, MovieLens\footnote{https://grouplens.org/datasets/movielens/1m/}. The characteristics of the MovieLens dataset are summarized in Table \ref{tab:movielens}. We use the version containing one million ratings, where each user has at least 20 ratings. We train an NCF framework with similar MLP network proposed in \cite{he2017neural} and followed their training strategies with Adam optimizer but without any pre-training. We test the \textit{predictive factors} of [8,16,32,64], and set the number of hidden units to the embedding size $\times 4$ in each hidden layer. We calculate both metrics for each test user and report the average score. For each method, we perform a wide range grid search of hyper-parameter $\lambda$ from $10^{-\alpha}$ where $\alpha \in {5,6,7,8,9}$ and report the experimental results based on the best performance on the validation set. The performance of a ranked list is judged by \textit{Hit Ratio} (HR) and \textit{Normalized Discounted Cumulative Gain} (NDCG) \cite{he2015trirank}. 

\begin{table}[ht]
	\vspace{-0.5em}
	\renewcommand\arraystretch{1.1}
	\setlength{\tabcolsep}{4pt}
	\caption{Statistics of the MovieLens datasets.}
	\begin{center}
	\small
	\begin{tabular}{|c|c|c|c|c|}
		\hline			
		\textbf{Dataset} & \textbf{Interaction\#} & \textbf{Item\#} & \textbf{User\#} & \textbf{Sparsity} \\
		\hline
		MovieLens & 1,000,209 & 3,706 & 6,040 & 95.53\%\\		
		\hline  
	\end{tabular}
	\end{center}
	\label{tab:movielens}
\end{table}

Figure \ref{fig:movielens} (a) and (b) show the performance of HR@10 and NDCG@10 w.r.t. the number of predictive factors. From this figure, it's clear to see that basis-path regularization achieve better generalization performance than all baseline methods. Figure \ref{fig:movielens} (c) and (d) show the performance of Top-K recommended lists where the ranking position K ranges from 1 to 10. As can be seen, the basis-path regularization demonstrates consistent improvement over other methods across positions, which is consistent with our analysis of generalization error bound in the previous section.

\subsection{Image Classification}

In this section, we apply our basis-path regularization to this task and conduct experimental studies based on CIFAR-10 \cite{cifar10}, with 10 classes of images. We employ a popular deep convolutional ReLU model, ResNet \cite{resnet} for image classification since it achieves huge successes in many image related tasks. In addition, we conduct our studies on a stacked deep CNN described in \cite{resnet} (refer to PlainNet), which suffers serious dependency among the paths. We train 34 layers ResNet and PlainNet networks on this dataset, and use SGD with widely used $l_2$ weight decay regularization (WD) as our baseline. In addition, we implement Q-SGD, which is proposed in \cite{ecsgd} and optimize the networks on basis paths. We investigate the combination of SGD/Q-SGD and basis-path regularization (BPR). Similar with the previous task, we perform a wide range grid search of $\lambda$ from $\{0.1,0.2,0.5\} \times 10^{-\alpha}$, where $\alpha \in \{3,4,5,6\}$. More training details can be found in supplementary materials.


\begin{table}[ht]
\renewcommand\arraystretch{1.1}
\setlength{\tabcolsep}{4pt}
\caption{Classification error rate (\%) on image classification task. Baseline is from \cite{resnet}, and the number of $\dagger$ is 7.51 reported in the original paper. Fig.~\ref{fig:plainresnet} shows the training procedures.}
\begin{center}
	\small
	\begin{tabular}{|c|c|c|c|c|c|c|}
		
		\hline  
		\multirow{2}*{Algorithm} & \multicolumn{3}{c|}{PlainNet34} & \multicolumn{3}{c|}{ResNet34} \\ \cline{2-7}  
		&  Train  &  Test  & $\Delta$  &Train  &  Test & $\Delta$ \\     
		\hline  
			
		SGD & 0.06 & 7.76&7.70&0.01&7.13&7.12\\
		SGD + WD & 0.06 & 6.34 &6.27&0.01&5.71$^\dagger$&5.70\\
		SGD + BPR & 0.06 & \textbf{5.99} &\textbf{5.92}&0.01&\textbf{5.62}&\textbf{5.61}\\
		\hline
		Q-SGD & 0.03 & 7.00 &6.97&0.01&6.66&6.65\\
		Q-SGD + BPR & 0.05 & \textbf{5.73} &\textbf{5.68}&0.03&\textbf{5.36}&\textbf{5.33}\\
		\hline  
	\end{tabular}
\end{center}
	\vspace{-0.em}
\label{tab:cifar10}
\end{table}

Table \ref{tab:cifar10} shows the training and test results of each algorithms. From the figure and table, we can see that our basis-path regularization indeed improves test accuracy of PlainNet34 and Resnet34 by nearly 1.8\% and 1.5\% respectively. Moreover, the training behaviors of SGD with weight decay and basis-path regularization are quite similar, but the basis-path regularization can always find better generalization points during optimization, which is consistent with our theoretical analysis in the previous section. We further investigate the combination of Q-SGD and basis-path regularization. Q-SGD with basis-path regularization achieves the best test accuracy on both PlainNet and ResNet model, which indicates that taking BP norm as the regularization term to the loss function is helpful for optimization algorithms.

\section{Conclusion}
In this paper, we define Basis-path norm on the group of basis paths, and prove that the generalization error of ReLU neural networks can be upper bounded by a function of BP norm. We then design Basis-path regularization method, which shows clearly performance gain on generalization ability. For future work, we plan to test basis-path regularization on larger networks and datasets. Furthermore, we are also interested in applying basis-path regularization on networks with different architecture.
	
\bibliography{arxiv}
\bibliographystyle{aaai}

\newpage

\section*{Supplementary: Capacity Control of ReLU Neural Networks by Basis-path Norm}

	This document contains supplementary theoretical materials and additional experimental details of the paper "Capacity Control of ReLU Neural Networks by Essential Path Norm".
\section{Skeleton Method to Construct Basis-Path}
We briefly review the construction of basis-paths. For an $L$-layer feedforward neural network with width $H$, skeleton methods \cite{ecsgd} construct basis paths following two steps:

(1) \textbf{Select skeleton weights:} The skeleton weights are the diagonal elements in weight matrix $w_l$, $l=1,\cdots,L-1$. For $w_0$, select elements $w_0(p_1 \mod d, p_1)$. For $w_L$, select elements $w_L(p_{L-1}, p_{L-1} \mod K)$. All the selected weights are called skeleton weights. Others are called non-skeleton weights. 

(2) \textbf{Construct basis-paths:} The paths contains no more than one non-skeleton weights are basis-paths.

In \cite{ecsgd}, the authors also proved the following properties of basis-paths: each non-skeleton weight will only appear in one basis-path.

\section{Proof of Proposition 1}
\begin{proposition}\label{p1}
	For any non-basis path $\tilde{p}$,  $\tilde{p}=\prod_{i=1}^{m}p_i^{\alpha_i}\prod_{j=m+1}^{r}p_j^{\alpha_j}$, where $\alpha_i\leq 0, \alpha_j\geq 0$.
\end{proposition}
\textbf{Proof:} We prove this proposition by induction. For ease of reference, we use $M_{i}(j)$ to denote the operator $i\mod j$, i.e., $M_{i}(j)=i\mod j$ where $i,j\in \mathbb{Z}_+$.

(1) If $L=2$, hidden node $O_{i_1}^1$ ($i_1=1,\cdots,H$) has an ingoing skeleton weight $w_1(M_{i_1}(d),i_1)$ and an outgoing skeleton weight $w_2(i_1,M_{i_1}(K))$. Then for a non-skeleton path $p_{(i_0,i_1,i_2)}$ where $i_0\neq M_{i_1}(d)$ and $i_2\neq M_{i_1}(K))$, it can be calculated as
\begin{align}
&p_{(i_0,i_1,i_2)}\\
=&w_1(i_0,i_1)\cdot w_2(i_1,i_2)\\
=&\frac{w_1(i_0,i_1)w_2(i_1,M_{i_1}(K))\cdot w_1(M_{i_1}(d),i_1)w_2(i_1,i_2)}{w_1(M_{i_1}(d),i_1)\cdot w_2(i_1,M_{i_1}(K))}\\
=&\frac{p_{(i_0,i_1,M_{i_1}(K))}\cdot p_{(M_{i_1}(d),i_1,i_2)}}{p_{(M_{i_1}(d),i_1,M_{i_1}(K))}}.
\end{align}According to skeleton method, $p_{(i_0,i_1,M_{i_1}(K))}$, $p_{(M_{i_1}(d),i_1,i_2)}$, $p_{(M_{i_1}(d),i_1,M_{i_1}(K))}$ are all basis-paths. We can see that the basis-path such as $p_{(M_{i_1}(d),i_1,M_{i_1}(K))}$ which contains no non-skeleton weights is in the denominator. The basis-path which contains one skeleton weight such as $p_{(i_0,i_1,M_{i_1}(K))}$ and $p_{(M_{i_1}(d),i_1,i_2)}$ is in the numerator. An example is shown in Figure 1 in the main paper.

(2) If the proposition is satisfied for a $L-1$-layer FNN, i.e.,
$$p_{(i_0,\cdots,i_{L-1})}=\prod_{l=1}^{L-1}w_l(i_{l-1},i_l)=\frac{\prod_{l\leq L-1}p_l(w_{i_{l-1},i_l})}{\prod_j p_j},$$where $p_l(w_l(i_{l-1},i_l))$ is the basis-path which contains non-skeleton weight $w_l(i_{l-1},i_l)$ \footnote{If $w_l(i_{l-1},i_l)$ is a non-skeleton weight. Otherwise, it will not appear in the numerator.} and $p_j$ denotes a basis-path which only contains skeleton weights. 

Then for a $L$-layer FNN, a non-basis path can be calculated as
\begin{align*}
&p_{(i_0,\cdots,i_{L-1},i_L)}\\=&\prod_{l=1}^{L}w_l(i_{l-1},i_l)\\
=&w_{L}(i_{L-1},i_L)\cdot\frac{\prod_{l\leq L-1}p_l(w_l(i_{l-1},i_l))\cdot w_L(w_l(i_{l-1},i_l))}{\prod_j p_j\cdot \prod_l w_L(w_l(i_{l-1},i_l))}\\
=&\frac{p_L(w_{L}(i_{L-1},i_L))}{\prod_{l=1}^{L-1}w_l(w_L(i_{L-1},i_L))}\cdot\frac{\prod_{l\leq L-1}p_l(w_l(i_{l-1},i_l))}{\prod_j p_j\cdot \prod_{l\leq L-1} w_L(w_l(i_{l-1},i_l))}\\
=&\frac{\prod_{l\leq L}p_l(w_l(i_{l-1},i_l))}{\prod_j \tilde p_j},
\end{align*}where $w_L(w_l(i_{l-1},i_l))$ is the skeleton weight at layer $L$ which connects the basis-path that contains $w_l(i_{l-1},i_l)$, and $w_l(w_L(i_{L-1},i_L))$ denotes the skeleton weight at layer $l$ that connects the weight $w_L(i_{L-1},i_L)$. Because $w_l(w_L(i_{L-1},i_L))$ are all skeleton weights, $\prod_{l=1}^{L-1}w_l(w_L(i_{L-1},i_L))$ is the value of a basis-path in a $L-1$ layer FNN. The establishing of the above equality also uses the fact that $p_l(w_l(i_{l-1},i_l))\cdot w_L(w_l(i_{l-1},i_l))$ is a basis-path for a L-layer FNN because it contains only one non-skeleton weight $w_l(i_{l-1},i_l)$. Combining $\prod_{l=1}^{L-1}w_l(w_L(i_{L-1},i_L))\cdot\prod_j p_j\cdot \prod_l w_L(w_l(i_{l-1},i_l))$ together, we can get that the denominator $\tilde p_j$ is a basis-path which contains only skeleton weight. 

Therefore, we prove that basis-path which contains one non-skeleton weight will only appear in the numerator and basis-path which only contains skeleton weights will only appear in the denominator in the calculation of the non-basis paths.

\begin{figure*}[htbp]
	\centering
	\includegraphics[width=1.0\textwidth]{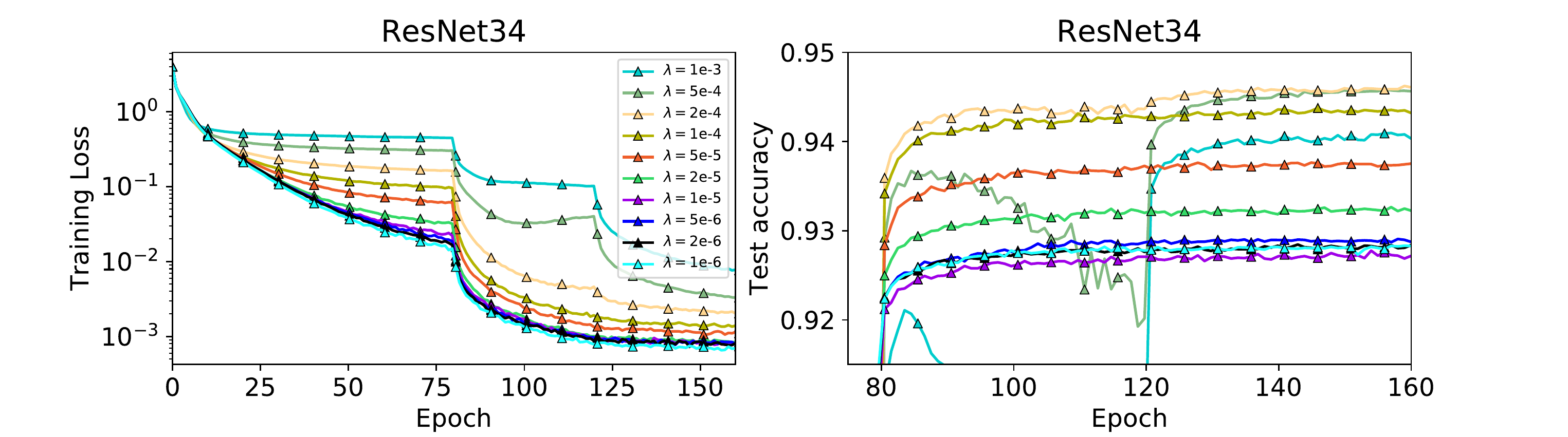}
	\caption{Training loss and test accuracy of ResNet34 models w.r.t. number of effective passes on CIFAR-10.}\label{fig:plainresnet_labmda}
\end{figure*}
\section{Experiments - additional material}

All experiments were conducted with \emph{Pytorch} commit 2b47480. The ResNet implementation can be found in (\emph{https://github.com/pytorch/vision/}). Unless noted the initialization method was used by sampling values from a uniform distribution with range $[-a,a]$, where $a=\frac{1}{\sqrt{h_{l-1}}}$, $h_{l-1}$ is the dimension of previous layer, which is the default initialization method for linear layers and convolutional layers in PyTorch.

\subsection{Experiment settings}
\subsubsection{Empirical Verification}
Multi-layer perceptrons with ReLU activation were trained by SGD in this experiments. The max iteration number is 500 epochs. The initial learning rate is set to 0.05 for each model. Exponential decay is a wildly used method when training neural network models\cite{hinton2012improving,ruder2016overview}, which is applied to the learning rate in this experiments with power 0.01. Mini-batch size with 128 was performed in this experiments.

\subsubsection{Recommendation System}
For the recommendation system experiments, to evaluate the performance of item recommendation, we employed the \emph{leave-one-out} evaluation, which has been widely used in literatures \cite{bayer2017generic,he2017neural}. For each user, we held-out the latest interaction as the test set and utilized the remaining data for training. We followed the common strategy that randomly samples 100 items that are not interacted by the user, ranking the test item among the 100 items.

\subsubsection{Image Classification}
For the image classification experiment, we use the original RGB image of CIFAR-10 dataset with size 3 $\times$ 32 $\times$ 32. As before, we re-scale each pixel value to the interval [0, 1]. We then extract random crops (and their horizontal	flips) of size 3 $\times$ 28 $\times$ 28 pixels and present these to the network in mini-batches of size 128. The training and test loss and the test error are only computed from the center patch (3 $\times$ 28 $\times$ 28).

We trained 34 layers ResNet and PlainNet models (refer to resnet34 and plain34 in the original paper respectively) on this dataset. We performed training for 64k iterations, with a mini-batch size of 128, and an initial learning rate of 1.0 which was divided by 10 at 32k and 48k iterations following the practice in the original paper. 

\subsection{Experimental Results on the Influence of $\lambda$}

As for different model and norm, the $\lambda$ should be selected carefully. As described in image classification task section in the main paper, we performed a wide range of grid search for each norm from $\{0.1,0.2,0.5\} \times 10^{-\alpha}$, where $\alpha \in \{3,4,5,6\}$, and reported the best performance based on the validation set. In this section, we show how $\lambda$ affect our basis-path regularization. The results are given in Figure \ref{fig:plainresnet_labmda}. Each result is reported by 5 independent runs with random initialization. Please note that too large value of $\lambda$ ($\lambda > 1e-3$ in this setting) will lead to diverge, meanwhile too small will make the regularization influence nearly disappear. A proper $\lambda$ will lead to significant better accuracy.

\end{document}